\documentclass{article}

\usepackage[english]{babel}

\usepackage[letterpaper,top=2cm,bottom=2cm,left=3cm,right=3cm,marginparwidth=1.75cm]{geometry}

\usepackage{amsmath}
\usepackage{amsfonts}
\usepackage{booktabs}
\usepackage{graphicx}
\usepackage[colorlinks=true, allcolors=blue]{hyperref}
\usepackage{svg}
\usepackage{subcaption}
\usepackage{graphicx}
\usepackage{algorithm2e}
\usepackage{xcolor}
\usepackage{enumerate}
\usepackage{natbib}
\usepackage{wrapfig}

\title{On The Planning Abilities of OpenAI's o1 Models:\\ Feasibility, Optimality, and Generalizability}
\author{Kevin Wang$^\nabla$, Junbo Li$^\nabla$, Neel P. Bhatt$^\nabla$, Yihan Xi, \\ Qiang Liu, Ufuk Topcu, and Zhangyang Wang$^\dag$
\thanks{All authors are with the University of Texas at Austin, Austin, TX, USA. ${}^\nabla$Equal contribution and co-first authors; ${}^\dag$Corresponding author; email: {\tt\small atlaswang@utexas.edu}.}%
}
\date{}

\begin{document}
\maketitle

\begin{abstract}
Recent advancements in Large Language Models (LLMs) have showcased their ability to perform complex reasoning tasks, but their effectiveness in planning remains underexplored. In this study, we evaluate the planning capabilities of OpenAI's o1 models across a variety of benchmark tasks, focusing on three key aspects: feasibility, optimality, and generalizability. Through empirical evaluations on constraint-heavy tasks (e.g., \textit{Barman}, \textit{Tyreworld}) and spatially complex environments (e.g., \textit{Termes}, \textit{Floortile}), we highlight o1-preview’s strengths in self-evaluation and constraint-following, while also identifying bottlenecks in decision-making and memory management, particularly in tasks requiring robust spatial reasoning. Our results reveal that o1-preview outperforms GPT-4 in adhering to task constraints and managing state transitions in structured environments. However, the model often generates suboptimal solutions with redundant actions and struggles to generalize effectively in spatially complex tasks. This pilot study provides foundational insights into the planning limitations of LLMs, offering key directions for future research on improving memory management, decision-making, and generalization in LLM-based planning. Code available at: \url{https://github.com/VITA-Group/o1-planning}

\end{abstract}


\section{Introduction}
\begin{wrapfigure}{r}{0.55\textwidth}
\vspace{-2.5em}
    \begin{center}
        \includegraphics[width=0.9\linewidth]{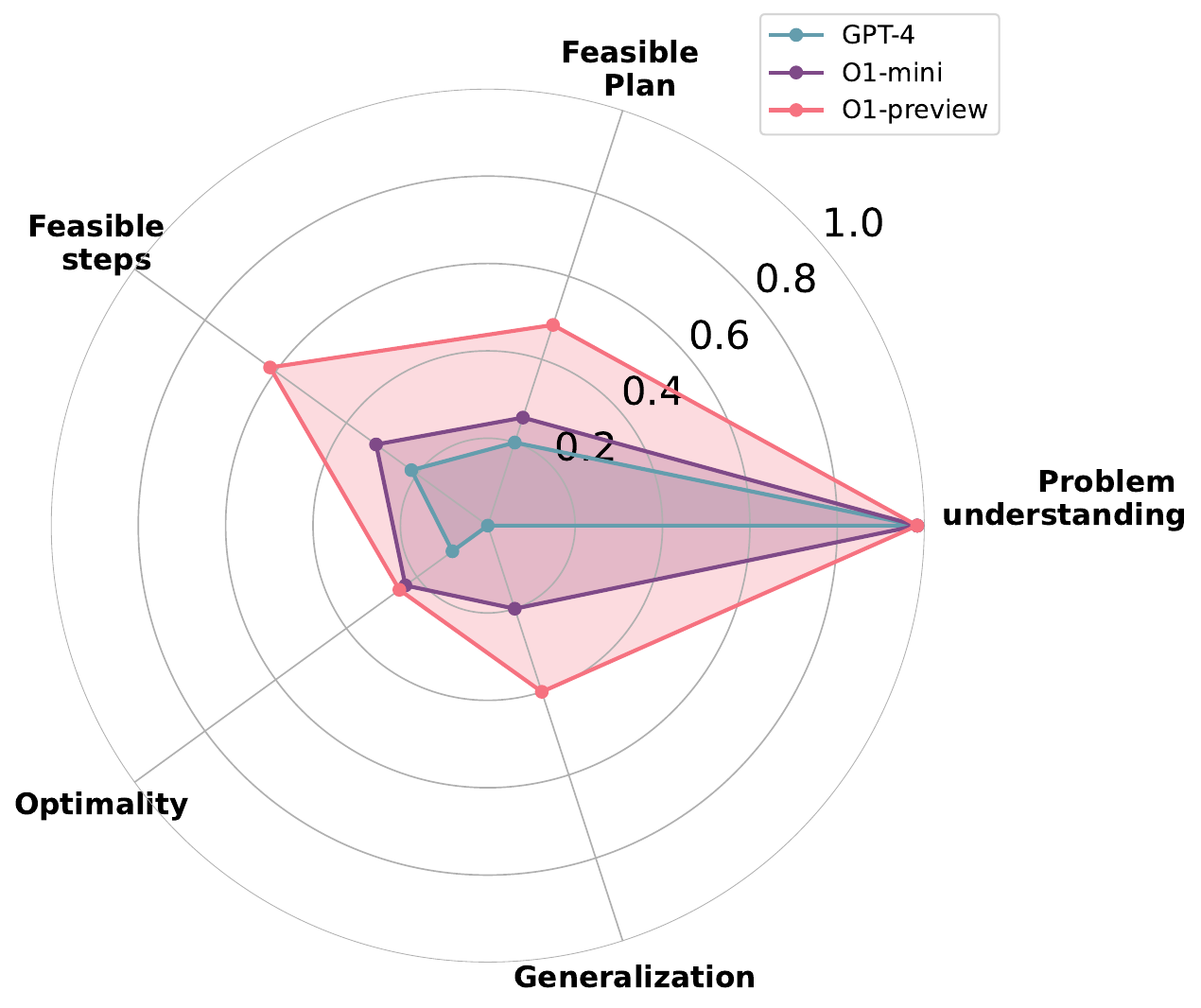}
    \end{center}\vspace{-1em}
    \caption{ Overall comparison of GPT-4, o1-mini,and o1-preview, on key planning perspectives defined by us.}\vspace{-1em}
    \label{fig:radar_chart}
\end{wrapfigure}

Large Language Models (LLMs) have significantly changed the landscape of artificial intelligence, achieving impressive results in various language-related tasks, such as chatbots, math, and coding, etc. One of the areas, that remains yet to be fully claimed by LLMs, is the use of language agents for planning in the interactive physical world \citep{huang2022inner, huang2022language, singh2023progprompt, lin2023text2motion}.
Previous scrutiny \citep{liu2023llm+,valmeekam2024llms,valmeekam2023planning,valmeekam2024planbench} pointed out that despite possessing advanced inference reasoning techniques like Chain-of-Thought (CoT) \citep{wei2022chain} and Tree-of-Thought (ToT) \citep{yao2024tree}, LLMs still struggle in making success plans without relying on external tools, such as a PDDL planner \citep{liu2023llm+, lyu2023faithful}.

The recent release of OpenAI's o1 model \citep{openai2024learning}, trained with reinforcement learning to naturally employ chain-of-thought reasoning, has reached new heights in problem-solving, particularly in mathematics and code generation. This suggests potential for planning, a seemingly related area. Recent research \citep{valmeekam2024llms} was the first to evaluate the success rates of o1 and other LLMs on \textit{Blocksworld} and its variants from PlanBench \citep{valmeekam2024planbench}, demonstrating that o1 has significantly enhanced its capabilities, extending the boundaries of what LLMs can accomplish in planning tasks.
Despite these improvements, o1 is still far from perfect, raising our curious question: \textit{where does the o1 model still fall short, and how can one systematically identify its limitations?}

Our work takes a deeper dive into the planning process to perform a more detailed analysis across a broader range of planning tasks building on \citep{valmeekam2024llms}. Rather than merely measuring the plan success rate as previous studies did \citep{valmeekam2024llms, liu2023llm+}, we classify the different types of errors LLMs make during their planning. Through extensive empirical evaluations, we analyze o1's performance across various domains from three key perspectives: \textit{plan feasibility, plan optimality}, and \textit{plan generalizability}. We aim to offer a clearer understanding of the current limitations of LLMs in the entire planning pipeline, and to facilitate future finer-grained diagnostics.

\section{Planning Ability Evaluation: Three Perspectives}\label{sec:Overview}
\label{sec:ability overview}

We propose evaluating the planning abilities of language model agents from three key perspectives: \textbf{feasibility, optimality}, and \textbf{generalizability}. These categories represent distinct but interconnected aspects of planning, each addressing a critical dimension of the agent's capability to effectively handle tasks in diverse environments. By dividing planning abilities into these three perspectives, we ensure a comprehensive evaluation, where each aspect plays a significant role in overall performance. The motivation for this division lies in the varied challenges planning entails, from basic execution to advanced optimization and adaptation across new contexts.

\paragraph{Feasibility} Feasibility assesses whether the agent can produce a viable plan to achieve the goal, often referred to as success rate in previous works \citep{liu2023llm+}. A plan must not only be executable but also ensure goal completion under real-world constraints. This category directly measures the agent's ability to operate within the problem’s rules, ensuring that each generated plan is valid and practical in real-world settings. Feasibility can be further divided into three key components:

\begin{enumerate}
    \item \textbf{Ability to create feasible steps} \quad Each step in a plan must be executable within the system, adhering to constraints specific to the problem domain. Constraints might include physical limitations, action order requirements, or other domain-specific rules. In route planning, for instance, certain zones may be inaccessible, while in operational planning, tasks may have dependencies that must be respected. We term failures related to this issue as ``Inability to Follow Problem Rules'' (\textbf{IR}). This occurs when a plan violates domain constraints, often because language models, trained on natural language data, misinterpret problem-specific constraints: some steps, although violating the physical constraints of the problem, may still seem logically plausible within a natural language framework, leading the agent to mistakenly adopt them. Such errors become more frequent as the complexity or number of rules increases, revealing a need for more sophisticated validation mechanisms beyond natural language reasoning.

    \item \textbf{Ability to generate a feasible plan} \quad Even if individual steps are valid, the overall plan may still fail to achieve the intended goal. The agent might not generate a coherent sequence of actions, leading to dead ends or random exploration. This issue, termed ``Inability to Generate a Feasible Plan'' (\textbf{IP}), grows more prominent in complex tasks requiring advanced reasoning. Stronger models like o1, which demonstrate superior reasoning capabilities, tend to perform better, as they provide more thorough analysis and structured plans.

    \item \textbf{Ability to understand the problem} \quad Feasibility also hinges on correctly interpreting the problem's initial and goal states. Even with valid steps and an overall plan, misinterpreting the starting conditions or the desired end state can result in errors. Such failures, termed ``Misinterpretation of Goal State'' (\textbf{MG}), are common when plans require deep reasoning over multiple steps. This issue, however, can be mitigated in models with stronger reasoning abilities.
\end{enumerate}

\paragraph{Optimality} While feasibility ensures that a plan can be successfully executed, optimality concerns how efficiently the plan achieves its goal. In many real-world scenarios, a feasible plan is not enough; the plan must also be resource-efficient, minimizing unnecessary actions, time, and cost. In this context, optimality refers to whether the language agent can generate the most efficient plan, avoiding redundant steps or suboptimal decisions that waste resources.

Optimality is particularly important in tasks with limited resources or strict time constraints, where even small inefficiencies can lead to significant performance degradation. For example, in a robotic task, an optimal plan would minimize the number of movements or tool changes required to complete the task, whereas a suboptimal plan might include unnecessary repetitions or idle steps. Failures in optimality, termed ``Lack of Optimality" (\textbf{LO}), arise when a plan, although feasible, includes extraneous or inefficient actions that prevent it from being considered the best solution.

Our experiments suggest that while advanced models like o1-preview show some improvements in generating more optimal plans compared to simpler models, the gap between feasibility and true optimality remains significant. This highlights the inherent difficulty for language models to reason not only about what needs to be done but also about how to do it in the most efficient manner. Developing strategies to incorporate cost-sensitive reasoning or employing more advanced search and pruning techniques could help mitigate this issue. Furthermore, improving optimality in LLMs requires better integration of domain-specific heuristics or optimization-based approaches that can help prioritize efficiency during the planning process.

\paragraph{Generalizability} Generalizability examines whether a language model can successfully plan across a diverse range of scenarios, including those it may not have explicitly encountered during training. Generalization becomes especially challenging when the agent must deal with abstract representations of actions or work in environments where the semantics of actions differ from those in natural language. This capability is crucial for robust performance in real-world applications, where the agent may encounter unfamiliar contexts or have to work with symbolic representations that are not directly tied to the real-world semantics - yet still following consistent logical structures.

Inspired by \citep{valmeekam2024planbench}, we test whether the agent can construct valid plans even when actions are represented by arbitrary symbols, devoid of any inherent natural language meaning. In these cases, the agent’s ability to generalize reflects its deeper understanding of the underlying structure and logic of planning tasks, independent of the specific linguistic cues it was trained on. This aspect is critical in fields like robotics, where planning often involves symbolic reasoning and manipulation of abstract entities. Our experiments (e.g., Figure \ref{fig:generalization results}) indicate that generalization remains a significant challenge for current models, especially in more complex, spatially dynamic environments. Models like o1-preview show a clear degradation in performance when transitioning from familiar tasks to generalized ones, suggesting that their learned representations are often too closely tied to specific task domains. This limitation highlights the need for more robust generalization mechanisms for LLM-based planners.

\begin{table}[htbp]
\centering
\resizebox{0.99\textwidth}{!}{%
\begin{tabular}{@{}l|lllll|lllll|lllllllll@{}}
\toprule
            & \multicolumn{5}{l|}{GPT-4}     & \multicolumn{5}{l|}{o1-mini}    & \multicolumn{5}{l}{o1-preview}  &  &  &  &  \\ \midrule
 &
  \multicolumn{1}{c}{\textbf{IP}} &
  \multicolumn{1}{c}{\textbf{LO}} &
  \multicolumn{1}{c}{\textbf{MG}} &
  \multicolumn{1}{c}{\textbf{IR}} &
  \multicolumn{1}{c|}{\textbf{SR}} &
  \multicolumn{1}{c}{\textbf{IP}} &
  \multicolumn{1}{c}{\textbf{LO}} &
  \multicolumn{1}{c}{\textbf{MG}} &
  \multicolumn{1}{c}{\textbf{IR}} &
  \multicolumn{1}{c|}{\textbf{SR}} &
  \multicolumn{1}{c}{\textbf{IP}} &
  \multicolumn{1}{c}{\textbf{LO}} &
  \multicolumn{1}{c}{\textbf{MG}} &
  \multicolumn{1}{c}{\textbf{IR}} &
  \textbf{SR} &
  \multicolumn{1}{c}{\textbf{}} &
   &
   &
   \\
Barman      & 0 & 0    & 0 & 10 & \textbf{0} & 0 & 0    & 0 & 10  & \textbf{0} & 0 & 0   & 0 & 10 & \textbf{0}   &  &  &  &  \\
Blocksworkd & 0 & 0   & 0 & 6  & 0.4        & 0 & 0   & 0 & 4   & 0.6        & 0 & 1   & 0 & 0  & \textbf{1}   &  &  &  &  \\
Floortile   & 0 & 0   & 1 & 9  & \textbf{0} & 1 & 0    & 0 & 9   & \textbf{0} & 7 & 0    & 0 & 3  & \textbf{0}   &  &  &  &  \\
Grippers    & 0 & 8   & 0 & 3  & 0.7        & 1 & 2   & 1 & 0   & 0.8        & 0 & 2   & 1 & 0  & \textbf{0.9} &  &  &  &  \\
Tyreworld   & 0 & N/A   & 0 & 9  & 0.1        & 0 & N/A  & 0 & 8 & 0.2          & 0 & N/A   & 0 & 0  & \textbf{1}   &  &  &  &  \\
Termes      & 0 & N/A & 0 & 10 & \textbf{0} & 0 & N/A & 0 & 10  & \textbf{0} & 0 & N/A & 0 & 10 & \textbf{0}   &  &  &  &  \\ \bottomrule
\end{tabular}%
}
\caption{The counts of each error type (IP, LO, MG, IR), as defined in Section~\ref{sec:Overview}. Note that LO only counts if the model outputs a feasible plan yet is not optimal. Besides, SR is the success rate (or feasible plan rate), with the highest success rate for each domain in bold. o1-preview has the highest success rate across all the domains.}
\label{table:results}
\end{table}

\section{Experiments on Planning Benchmarks}

\begin{figure}[hbt!]
    \centering
    \includegraphics[width=1.0\linewidth]{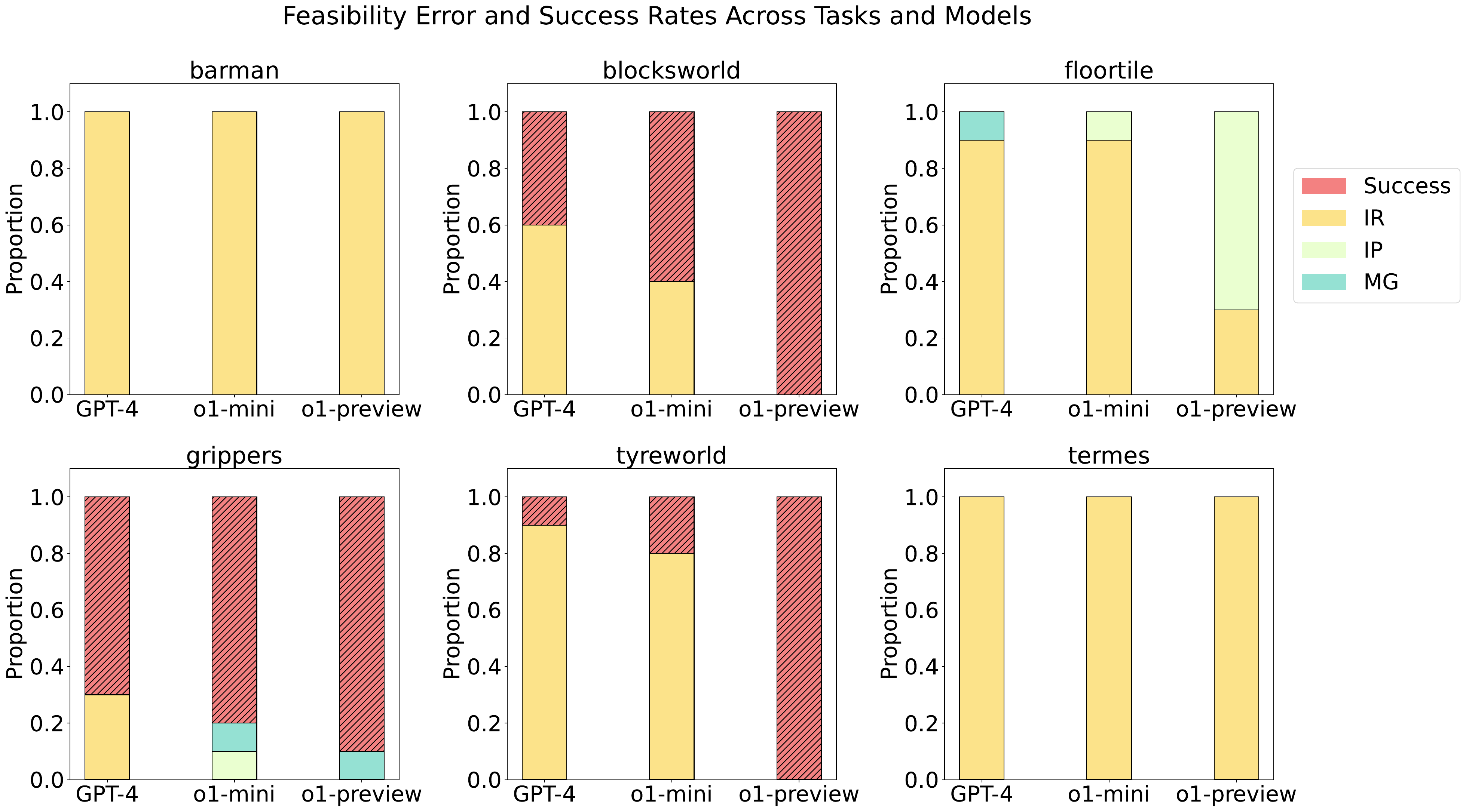}
    \caption{Feasibility error and success rate for 6 tasks and 3 models. Overall, o1 improves the success rate for certain tasks, but many problematic issues still persist. Examples of different error types are detailedd in later figures:  IR: \ref{fig:barman fail}, \ref{fig:blocksworld_1}, \ref{fig:floortile1}, \ref{fig:termes}, \ref{fig:tyrewrold} ; IP: \ref{fig:floortile1} ; MG: \ref{fig:gripper1}}.
    \label{fig:feasibility results}
\end{figure}

\begin{figure}[hbt!]
    \centering
    \includegraphics[width=0.7\linewidth]{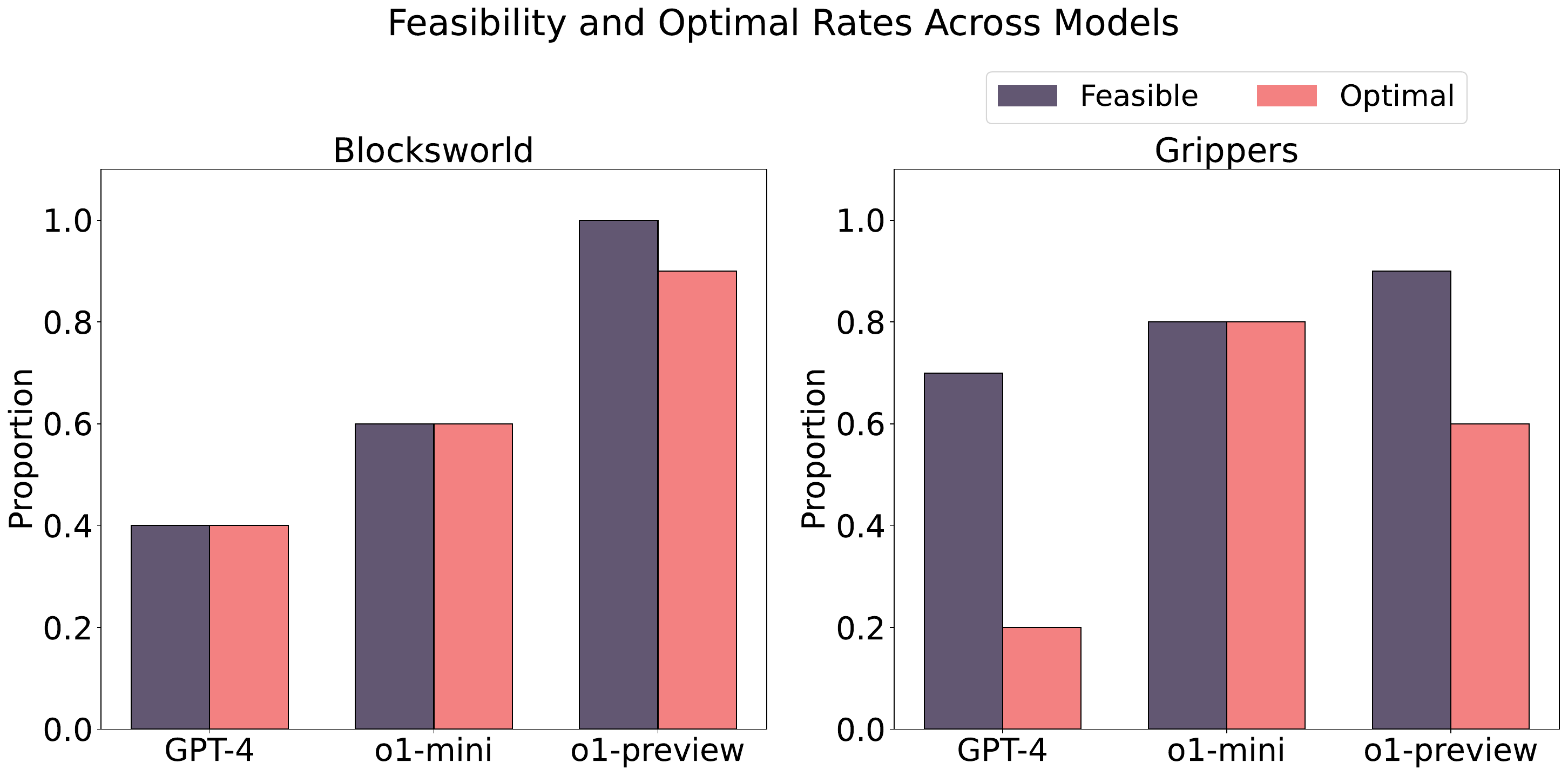}
    \caption{Success rate and optimality rate for \textit{Blocksworld} and \textit{Grippers}. Compared to GPT-4, o1 can provide more optimal plans. Example of suboptimal solutions are provided in Figures \ref{fig:blocksworld_2} and \ref{fig:gripper2}. }
    \label{fig:optimality results}
\end{figure}

\begin{figure}[hbt!]
    \centering
    \includegraphics[width=0.7\linewidth]{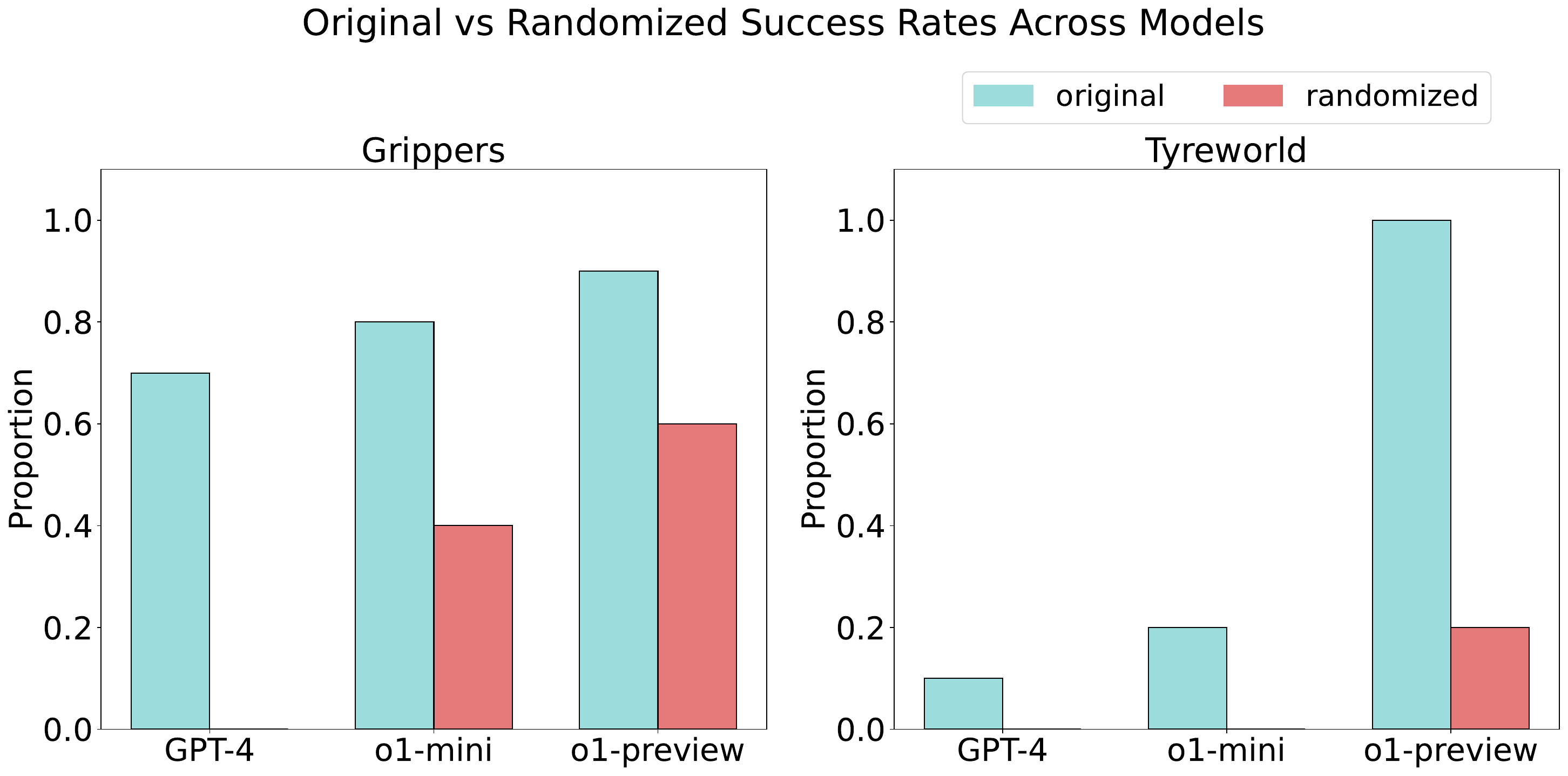}
    \caption{Success rate for generalization setting. GPT-4 fails entirely on challenging generalized tasks, whereas o1 is able to solve some of them. An example of randomize domain is provided in Figure \ref{fig:randomized_typreworld}.}
    \label{fig:generalization results}
\end{figure}

We assess the planning capabilities of GPT-4, o1-mini, and o1-preview as discussed in Section \ref{sec:ability overview}. The overall comparison in outlined in Table \ref{table:results}. Next, we will explore each task in detail to carefully analyze the language agents' abilities across various planning tasks. The results afor feasibility, optimality, and generalizability are presented in figures \ref{fig:feasibility results}, \ref{fig:optimality results}, and \ref{fig:generalization results}, respectively. For each task, errors are highlighted in red, while the reasons for these errors are indicated in orange. The specific action constraints that are violated are discussed in the accompanying captions. 
Details about the action domain and additional examples for each problem are available at: \url{https://github.com/VITA-Group/o1-planning}.

\subsection{Barman}

\paragraph{Task description}
In this task, a robot barman is tasked with preparing a series of drinks by manipulating drink dispensers, shot glasses, and a shaker. The robot, equipped with two hands, must perform a variety of actions such as grasping containers, filling/refilling shot glasses, pouring ingredients, shaking cocktails, and cleaning or emptying containers. Each action comes with strict preconditions—for example, the robot can only grasp a container when one hand is free, and shaking a cocktail is only possible when the shaker contains exactly two ingredients. Successfully completing this task requires precise sequencing of actions, where adhering to these constraints is crucial to avoid mistakes.

\begin{figure}[hbt!]
    \centering
    \includegraphics[width=1.0\linewidth]{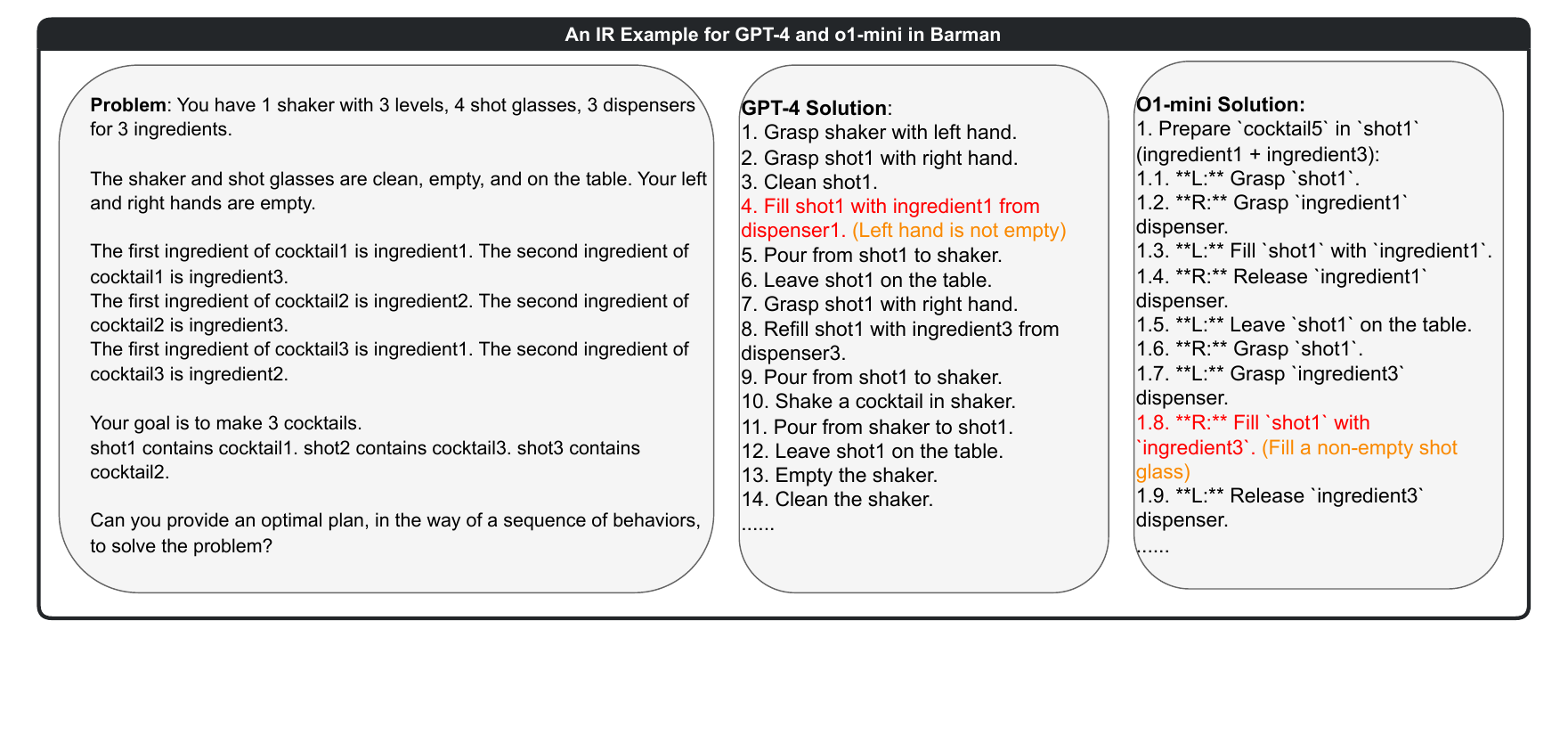}
    \vspace{-1.5cm}
    \caption{A failure example for Barman. The left side contains the problem statement, while the right side shows the first lines of the solutions provided by GPT-4 and o1-mini. The GPT-4 solution fails because the rules require that one hand must be empty for ``filling'', while the o1-mini solution fails because the rules specify that ``filling'' applies only to an empty shot glass.}
    \label{fig:barman fail}
\end{figure}

\paragraph{Analysis}
The results indicate that the language model agent struggles significantly with this task, consistently failing to generate feasible plans. Nearly all the errors stem from the agent’s inability to follow the specified rules, categorized as the IR error. For example, the rules demand that certain actions, like filling a container, require one hand to be free, or that specific actions must be taken in a strict order, such as holding a container before filling it. However, the LLM agents often overlook these critical constraints. This issue persists even in more advanced models like o1-preview. Figure \ref{fig:barman fail} illustrates the first error in the solution, where both GPT-4 and o1-mini violate explicit rules, highlighted in red and orange, respectively. These errors underline a common limitation of language models: while they can generate plausible sequences of actions in a natural language context, they frequently overlook key operational constraints critical for real-world execution.

\subsection{Blocksworld}

\begin{figure}[hbt!]
  \centering
    \centering
    \begin{subfigure}{0.49\textwidth}
    \includegraphics[width=\linewidth]{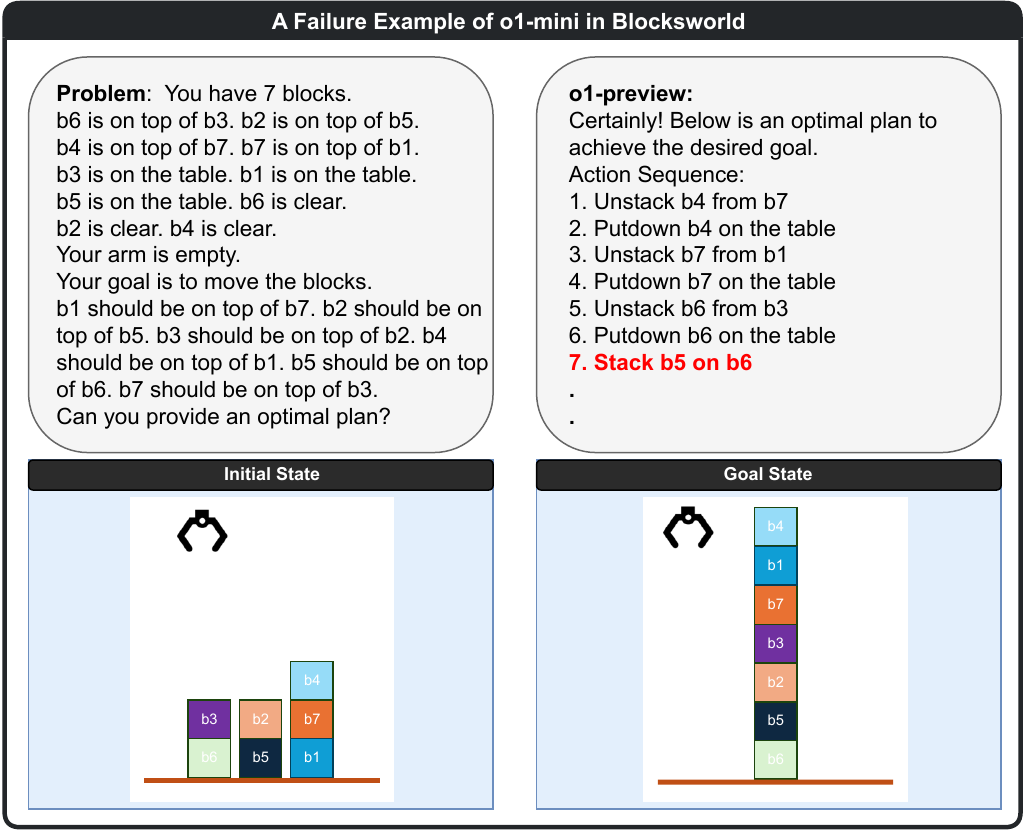}
    \caption{}
    \label{fig:blocksworld_1}
    \end{subfigure}
    \begin{subfigure}{0.49\textwidth}
    \includegraphics[width=\linewidth]{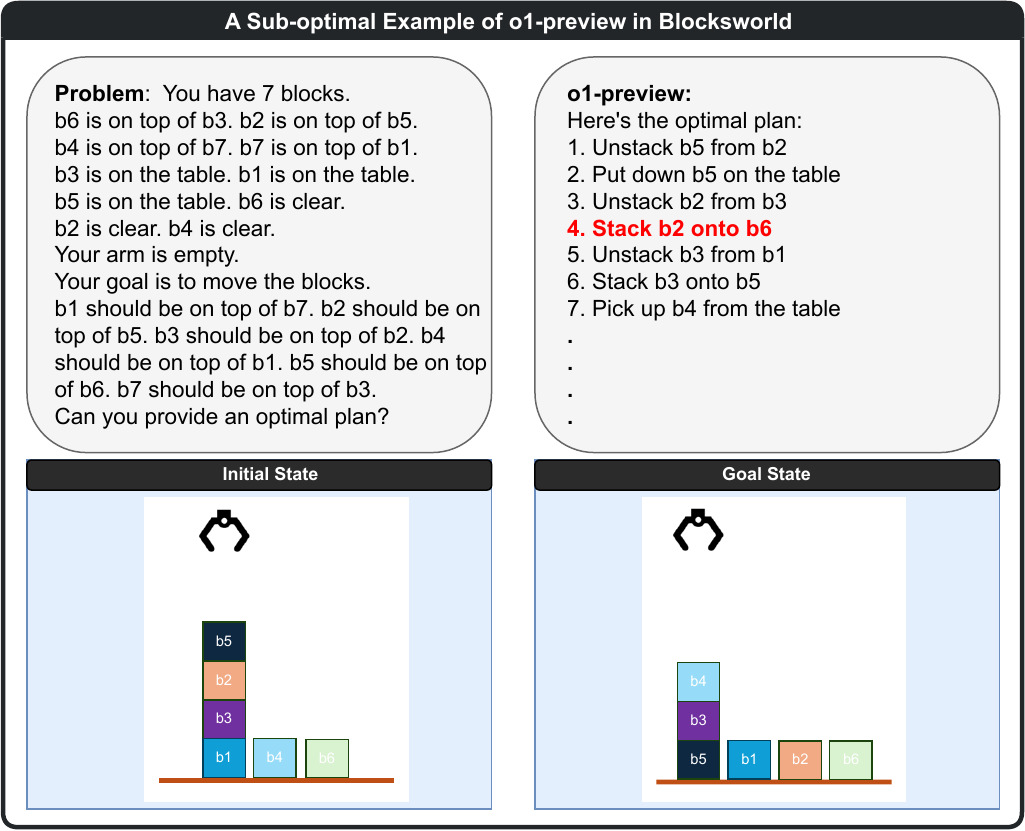}
    \caption{}
    \label{fig:blocksworld_2}
    \end{subfigure}
    \caption{Example failure and sub-optimal cases in the Blocksworld planning problem. (a) An illustration of a plan generated by o1-mini showcasing IR error in step 7. (b) An illustration of a sub-optimal plan generated by o1-preview showcasing LO error in step 4.}
    \label{fig:blocksworld}
\end{figure}

\paragraph{Task description} 
This planning task involves multiple blocks arranged on a table, where the goal is to move from an initial configuration to a pre-specified goal configuration. The robot arm, which can hold only one block at a time, must execute a series of actions such as picking up, putting down, stacking, and unstacking blocks to achieve the desired arrangement. The challenge lies in determining the correct sequence of these actions, while adhering to constraints that dictate how blocks can be manipulated. The action space is limited to fundamental operations like pickup, putdown, and stack.

\paragraph{Analysis} 
In this task, GPT-4 demonstrated a relatively low success rate of 40\%, while o1-mini performed slightly better at 60\%. However, o1-preview achieved a perfect 100\% success rate, reflecting its stronger reasoning capabilities. The success rates decreased as the number of blocks increased, highlighting the challenge of managing larger object sets. Both GPT-4 and o1-mini frequently failed to follow problem constraints, an issue categorized as the IR error. Figure \ref{fig:blocksworld_1} provides an example where o1-mini fails to comply with these constraints.

Although o1-preview successfully completed all tasks, it was not entirely optimal. In one instance, the model added an unnecessary step, leading to a suboptimal solution despite reaching the correct goal state. This issue, characterized as LO, is illustrated in Figure \ref{fig:blocksworld_2}. The occurrence of suboptimal steps even in successful models emphasizes the ongoing challenge of optimizing planning tasks, where generating feasible solutions is not always sufficient.

\subsection{Grippers}

\paragraph{Task description}
This task involves a team of robots equipped with two grippers, capable of moving between rooms and manipulating objects. The robots have three primary actions: moving from one room to another, picking up objects, and dropping them. Each action is constrained by the robot’s current location and the status of its grippers, meaning that a robot can only pick up an object if its gripper is free and can only drop an object in a specific location once it is carrying one. Effective planning requires coordinating these actions while adhering to these constraints to accomplish the goal of manipulating objects across different rooms.

\begin{figure}[hbt!]
  \centering
  \begin{subfigure}{0.49\textwidth}
    \centering
    \includegraphics[width=1\textwidth]{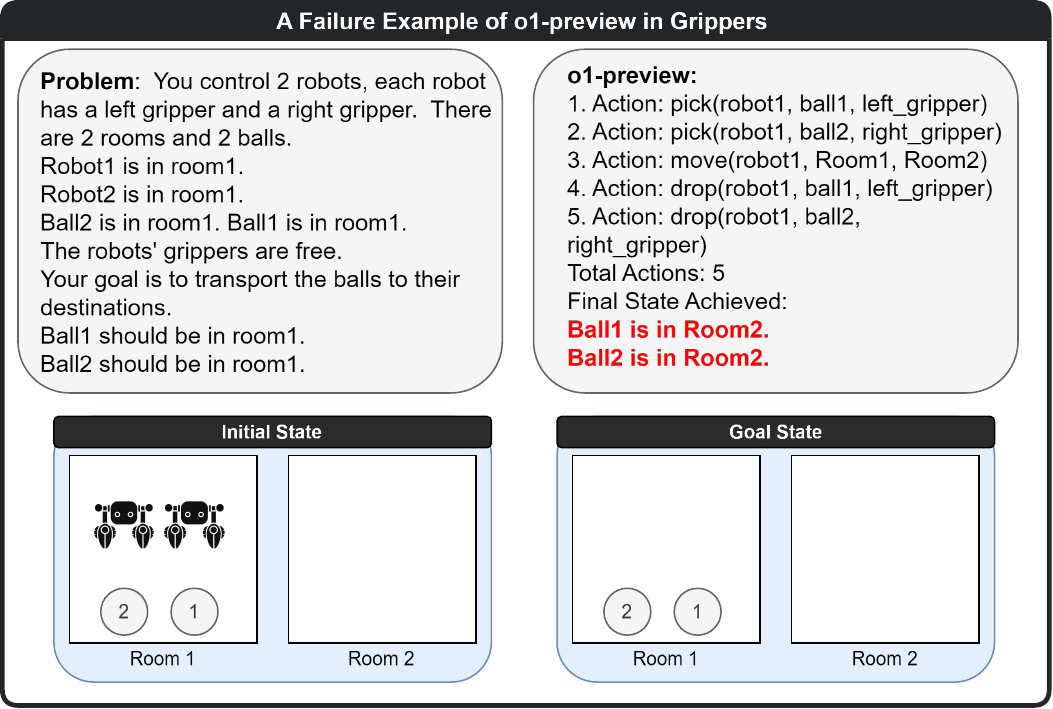}
    \caption{}
    \label{fig:gripper1}
  \end{subfigure}
  \hfill 
  \begin{subfigure}{0.49\textwidth}
    \centering
    \includegraphics[width=1\textwidth]{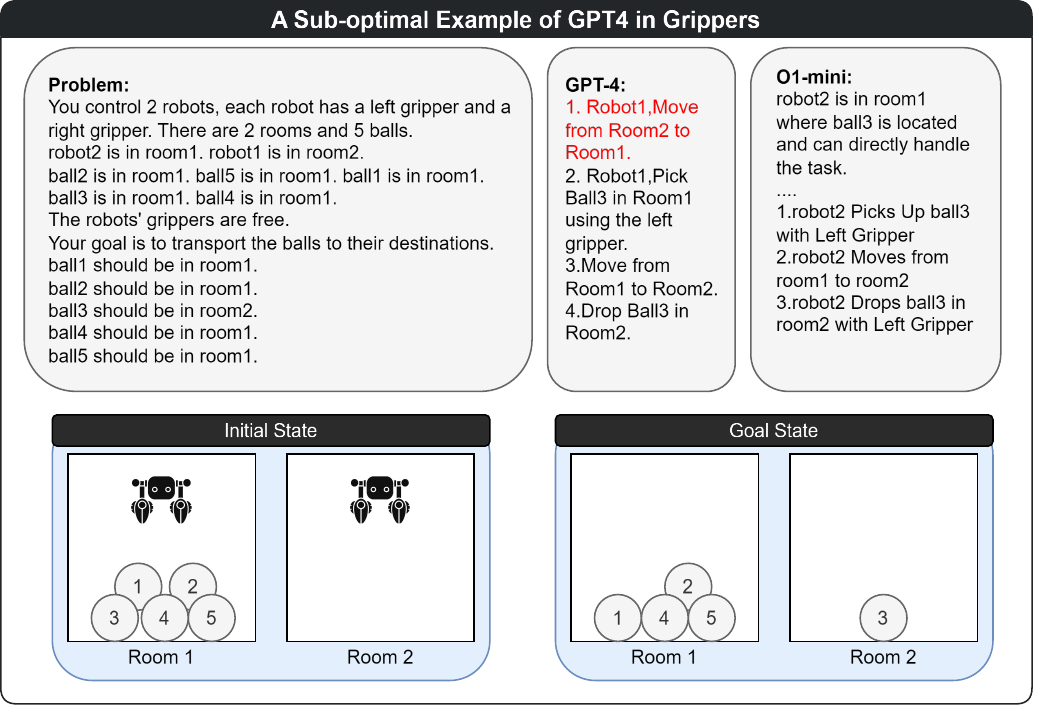}
    \caption{}

    \label{fig:gripper2}
  \end{subfigure}
  \caption{Failure examples in Grippers. (a) showcases o1-preview MG error: o1-preview assumes the goal state is both balls in room2 instead of room1; (b) highlights GPT4's suboptimality: it took an extra step to complete the goal, while o1-mini can return the optimal plan.}
  \label{fig:main}
\end{figure}

\paragraph{Analysis}
In this domain, both o1-mini and o1-preview significantly outperformed GPT-4, particularly in success and optimality rates. GPT-4 managed a 70\% success rate but only a 20\% optimality rate, indicating frequent suboptimal action sequences. In contrast, o1-mini achieved both higher success and optimality rates, at 80\% for each. o1-preview performed even better with a 90\% success rate, though its optimality rate dropped slightly to 70\%. Figure \ref{fig:gripper2} showcases an example where GPT-4 generates a suboptimal solution by including an unnecessary relocation of the robot, which adds redundant steps to the plan. Conversely, o1-mini, through additional reasoning, eliminated unnecessary movements, leading to an optimal solution.

However, o1-preview exhibited a shortcoming related to the MG error. In one instance, depicted in Figure \ref{fig:gripper1}, the robot's initial state already coincided with the goal state—both balls were already in the correct room (room1), meaning no action was required. While both GPT-4 and o1-mini correctly recognized that no further steps were necessary, o1-preview mistakenly identified room2 as the goal, leading to unnecessary actions aimed at relocating the balls. This misinterpretation demonstrates that even advanced models like o1-preview can struggle with maintaining accurate goal-state representations, especially in tasks requiring minimal or no action.

\subsection{Floortile}

\paragraph{Task description}
In this task, a team of robots is responsible for painting a grid of floor tiles in black and white. Each robot can move in four directions, switch the color of its spray gun, and paint tiles directly in front of or behind them. The main challenge is that robots can only paint tiles that are currently unpainted and cannot move onto tiles that have already been painted. This creates a complex constraint, requiring careful planning of movements and actions to achieve the desired tile pattern without the robots trapping themselves or each other. The task demands strategic coordination between movement and painting actions, ensuring the robots follow the rules while efficiently completing the grid pattern.

\paragraph{Analysis}
In this domain, all models—GPT-4, o1-mini, and o1-preview—failed to solve the test cases, but the reasons for their failures varied. For GPT-4 and o1-mini, 90\% of their failures stemmed from the IR error.  Specifically, both models frequently violated the rule that robots can only paint tiles directly in front or behind them, instead attempting to paint the tile on which they were standing. This rule violation was a common source of errors as the models struggled to keep track of the task constraints while moving and painting simultaneously.

\begin{figure}[hbt!]
    \centering
    \includegraphics[width=1.0\linewidth]{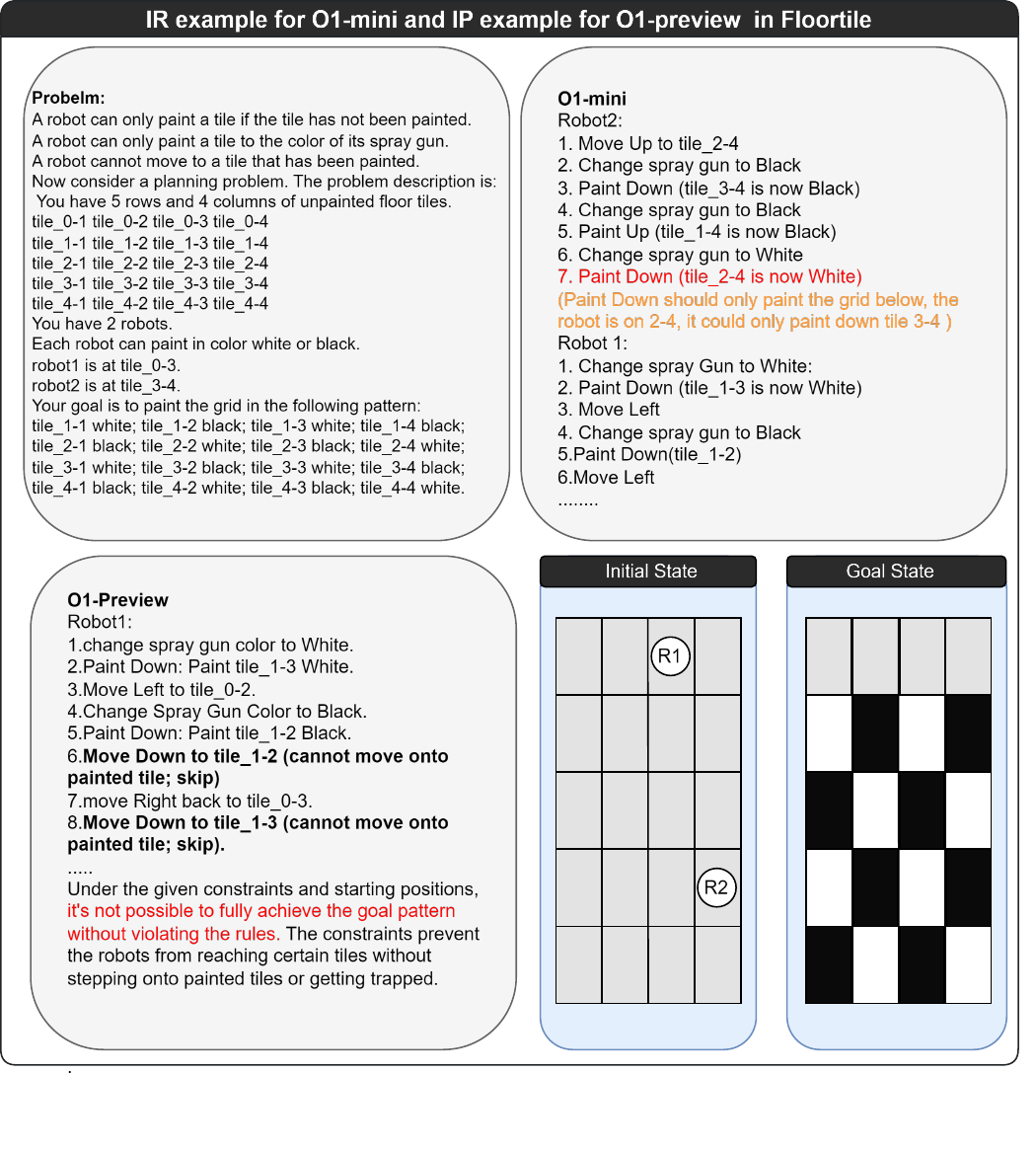}
    \vspace{-1.5cm}
    \caption{We illustrate the IR error of o1-mini (top right) and the IP error of o1-preview (bottom left). o1-mini incorrectly assumed that ``paint down" would paint the grid beneath the robot. Meanwhile, o1-preview adhered to the constraints throughout the plan but ultimately failed to complete it.}
\label{fig:floortile1}
\end{figure}

On the other hand, o1-preview showed a notable improvement in this regard, with only 30\% of its failures caused by IR. o1-preview's internal self-evaluation mechanism allowed it to better track the rules and adjust its actions accordingly. For instance, when it initially attempted to paint the wrong tile, it was able to reevaluate the action and correct itself by following the task constraints. However, despite these improvements in rule adherence, o1-preview encountered other errors, such as rule confusion. In some cases, it misinterpreted which tiles could be painted or made invalid assumptions about the sequence of movements. While its chain-of-thought reasoning helped it self-correct in some cases, it was ultimately unable to solve the task entirely, as seen in Figure \ref{fig:floortile1}.

\subsection{Termes}
\paragraph{Task description}
The Termes task requires controlling a robot to construct structures by moving between different positions and manipulating blocks. The robot can move horizontally, vertically (up and down), and is tasked with placing or removing blocks at neighboring positions that match in height. Additionally, the robot can create new blocks at a depot or destroy blocks when needed. To achieve the specified construction goals, the robot must efficiently plan its movements and strategically use blocks, adhering to height and positional constraints throughout the task. The challenge lies in coordinating these actions while respecting the rules that govern both movement and block placement.

\paragraph{Analysis}
All models—GPT-4, o1-mini, and o1-preview—failed to successfully complete the tasks in the Termes domain, largely due to shortcomings in detailed planning. A common source of error across all models was the failure to account for the task’s height constraints when moving horizontally, upward, or downward. These constraints, which ensure that the robot can only move to positions of matching or specific relative heights, were frequently ignored by the models. Additionally, the language models often made the mistake of placing blocks at their current position, a violation of the task’s rule that blocks can only be placed at neighboring positions.

\begin{figure}[hbt!]
    \centering
    \includegraphics[width=1.0\linewidth]{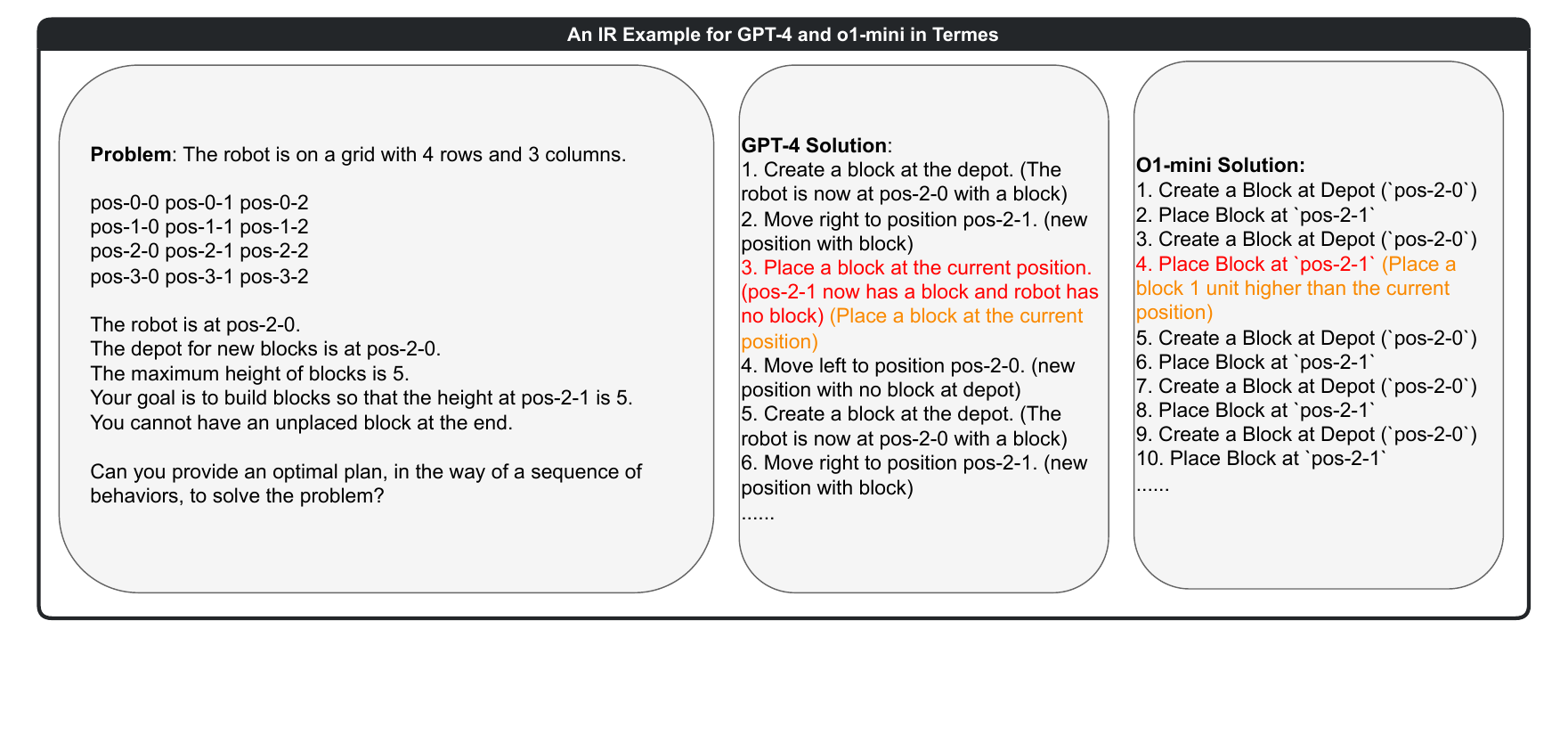}
    \vspace{-1.5cm}
    \caption{A failure example for Termes. GPT-4 solution fails because the block can only be placed in a neighboring position, while o1-mini solution fails since the block must be placed at the same height.}
    \label{fig:termes}
\end{figure}

These errors highlight the models' difficulty in managing complex spatial relationships and adhering to the intricate rules of the task. Although the actions generated by the models may seem plausible in a natural language context, they frequently overlook key operational details required for real-world robot planning. This inability to follow task-specific constraints leads to failures in execution, as illustrated in Figure \ref{fig:termes}, where the model skips necessary steps and misinterprets the correct sequence of actions.

\subsection{Tyreworld}

\paragraph{Task description}
This task involves replacing flat tyres on vehicle hubs with intact, inflated tyres. The process requires the use of tools such as a wrench, jack, and pump, and the agent must follow specific actions to manipulate the tyres, nuts, and tools. There are 11 predefined actions, including opening and closing the boot, fetching and storing tools, loosening and tightening nuts, jacking up and down hubs, removing and installing wheels, inflating tyres, and securing or undoing nuts. Success in this task depends on executing these actions in the correct sequence while satisfying specific preconditions, such as using the wrench to loosen the nuts before removing the wheel or tightening the nuts only after lowering the jack.

\begin{figure}[hbt!]
    \centering
    \includegraphics[width=1.0\linewidth]{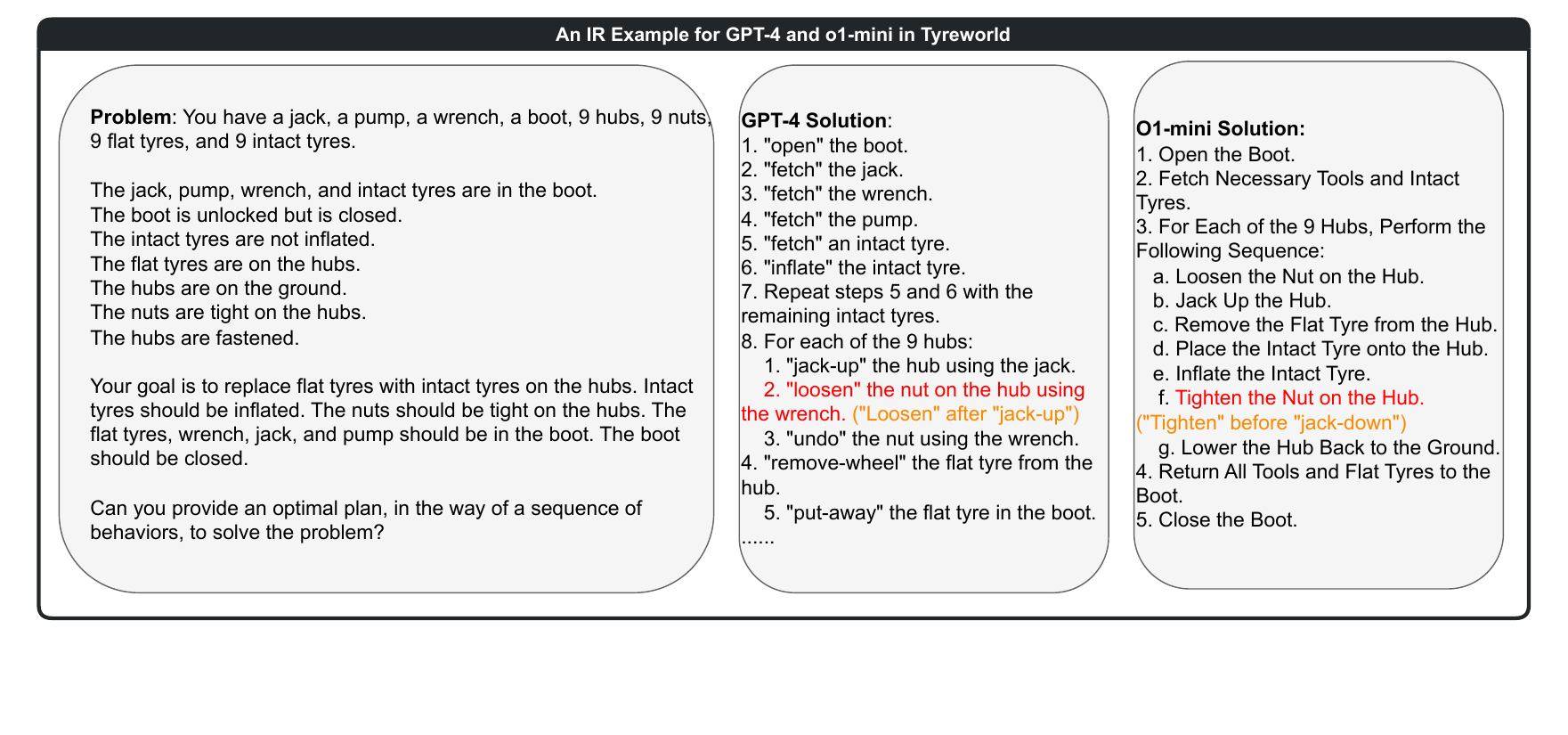}
    \vspace{-1.5cm}
    \caption{A failure example for Tyreworld. GPT-4 fails since ``loosen'' must occur before ``jack-up'', and o1-mini fails since ``tighten'' must happen after ``jack-down''.}
    \label{fig:tyrewrold}
\end{figure}

\paragraph{Analysis}
o1-preview generated correct plans for all test problems, significantly outperforming both GPT-4 and o1-mini, which failed to complete all but the simplest cases. The primary issue for GPT-4 and o1-mini was their frequent failure to follow the required action sequences. For instance, common errors included “loosening the nuts after jacking up” or “tightening the nuts before jacking down”—critical mistakes that would prevent successful tyre replacement. Figure \ref{fig:tyrewrold} provides an example  where failing to follow these mechanical constraints leads to incorrect plans.

\begin{figure}[hbt!]
    \centering
    \includegraphics[width=0.95\textwidth]{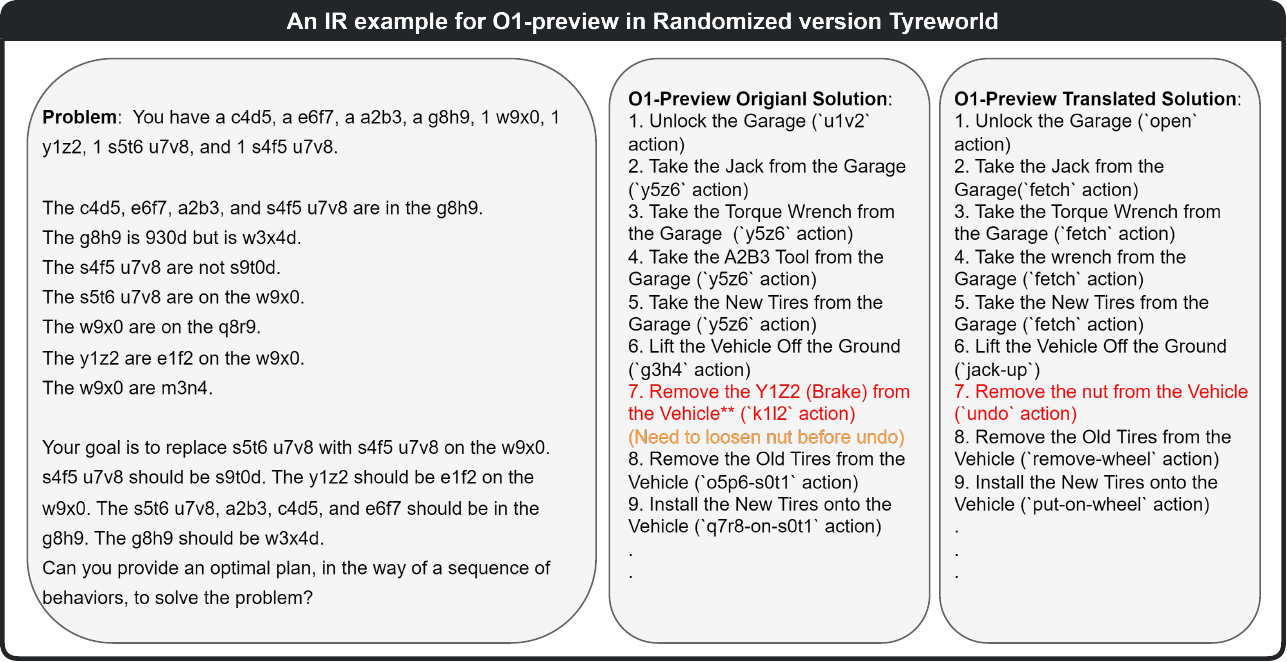}
    \caption{An example from the Randomized Tyre World: The center panel displays O-1's original solution, while the right panel translates the random symbols back to their original terms. Here, O-1 failed to adhere to the constraint that the nut must be loosened before it can be undone.}
    \label{fig:randomized_typreworld}
\end{figure}

We also evaluated the models’ generalization capabilities within this domain, revealing that while o1-preview performed well on structured tasks, its success rate dropped from 100\% to 20\% when the actions and tools were replaced with random symbols. Despite the symbols being abstract, the underlying action constraints and logic remained unchanged. This sharp decrease in performance, as illustrated in Figure \ref{fig:randomized_typreworld}, suggests that while o1-preview excels in rule-based planning with familiar symbols, it struggles to generalize when the problem context becomes more abstract.

\section{Discussion}
\label{sec:discussion}

\subsection{Empirical Limitations}
The primary limitation of this study stems from the relatively small dataset used in our empirical evaluations. While the experiments provide a foundational understanding of the o1 model’s planning capabilities, broader insights into its generalizability and robustness can only be derived with more extensive testing across larger and more diverse datasets. Larger datasets would help expose potential weaknesses that remain hidden in smaller, more structured environments, and would allow us to explore how o1 models handle a wider variety of constraints and complexity levels. Future work shall test in more real-world settings, where planning problems involve dynamic and less predictable elements.

\subsection{Model Performance vs. Problem Complexity}

\begin{wrapfigure}{r}{0.42\textwidth}
    \begin{center}
    \vspace{-1cm}
        \includegraphics[width=0.9\linewidth]{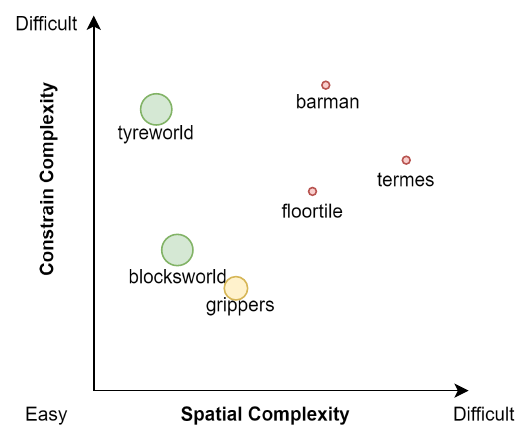}
    \end{center}
    \vspace{-0.6cm}
    \caption{We empirically evaluated these problems based on two complexity dimensions: action complexity and spatial complexity. The color coding represents the success rates of the o1-preview model: \textcolor{green}{green} indicates a high success rate, \textcolor{yellow}{yellow} moderate success, and \textcolor{red}{red} complete failure.}
    \vspace{-1cm}
    \label{fig:problem_complexity}
\end{wrapfigure}

Our analysis reveals a strong correlation between the complexity of the problem and the performance of the o1 model. We empirically examine each problem along two dimensions of complexity: action complexity and spatial complexity, as illustrated in Figure \ref{fig:problem_complexity}.

Specifically, the \textit{Floortile} and \textit{Termes} tasks highlight the challenges o1 faces in environments with higher spatial and rule-based complexity. In \textit{Floortile}, the task is set in a two-dimensional world, where robots must follow strict painting rules while navigating a constrained grid. In contrast, \textit{Termes} involves a three-dimensional setting, introducing additional layers of complexity due to vertical movement constraints and the need for precise block manipulation. Interestingly, the size of the action space did not appear to significantly affect the model’s ability to capture and use context efficiently. Instead, the complexity of spatial relationships and state transitions proved to be more problematic. This suggests that while o1 models can handle tasks with limited actions (e.g., \textit{Grippers}), they struggle when required to reason about more abstract, multi-dimensional spaces where maintaining an accurate internal state becomes critical.

\subsection{Constraint Following and State Management}
One key finding of this study is the o1 model’s improved ability to follow constraints and manage states, especially in comparison to GPT-4. o1-preview's self-evaluation mechanism, which allows the model to check and correct its actions during plan generation, was particularly effective in tasks like \textit{Blocksworld} and \textit{Tyreworld}. In these tasks, o1-preview demonstrated a higher success rate in adhering to complex rules, such as the preconditions for using a wrench or jack in \textit{Tyreworld}, while avoiding the rule violations that plagued GPT-4 and o1-mini. However, this ability to follow constraints deteriorates in more complex environments like \textit{Termes}, where the need for precise spatial reasoning and multi-step manipulation often leads to rule violations and misinterpretations of task goals. This points to a potential limitation in the model’s state management when dealing with more abstract problem spaces, and may call for explicit integration of neurosymbolic methods \cite{yang2024fine}.

\subsection{Optimality and Redundancy in Planning}
Optimality remains a significant challenge for the o1 models, as demonstrated across tasks like \textit{Blocksworld} and \textit{Floortile}. While o1-preview often generated feasible plans, it frequently failed to produce optimal solutions, leading to redundant actions and inefficiencies. For example, in \textit{Blocksworld}, o1-preview added unnecessary steps to the plan, reducing its overall efficiency despite reaching the correct goal state. This suggests that while the model can understand and follow constraints, it struggles with decision-making related to resource minimization and action optimization. The ability to reason about optimality is crucial for real-world applications, where minimizing steps and resources is often as important as achieving the correct outcome. Enhancing this aspect of o1’s reasoning mechanism—perhaps by incorporating more advanced cost-based decision frameworks—would be a valuable area for future research. Additionally, we observed that all three models exhibited some hallucination, including the assumption of non-existent rules. For instance, the o1-preview model in the grippers assumed that it could only move to adjacent numbered rooms, whereas the actual rule specifies that the move action can proceed to any room. Although model could still generate feasible plan, but that also hinder its ability to generate optimal plan.

\subsection{Generalization and Adaptability}
Another promising outcome of this study is o1-preview's demonstrated ability to generalize across tasks with consistent rule structures, as seen in \textit{Grippers}. In these cases, o1-preview outperformed GPT-4 by effectively adapting its learned strategies to new environments. In these scenarios, o1-preview consistently outperformed GPT-4 by effectively adapting its learned strategies to new environments. The o1 model attempted to imbue meaningless symbols with natural language meaning to aid problem-solving, as seen in Figure \ref{fig:randomized_typreworld}. Additionally, o1-preview's self-evaluation capabilities enabled it to maintain reasonable adherence to constraints, with only minor deviations, compared to GPT-4, which often fails to grasp the goal. While o1-preview's generalizability surpases previous GPT4 model, particularly in structured, low-dimensional tasks, there is still substantial room for improvement in enabling these models to adapt to more dynamic, high-dimensional, and abstract problem spaces.

\section{Conclusion}
\label{sec:conclusion}

Our study offers a pilot evaluation of the planning capabilities of OpenAI’s o1 models, providing new insights into their strengths and limitations. By systematically evaluating their feasibility, optimality, and generalizability across diverse planning tasks, we have uncovered key areas where o1-preview demonstrates promising advancements as well as significant challenges that remain to be addressed.

\subsection{Summary of Findings}
The findings of our experiments can be summarized from four key perspectives:
\begin{enumerate}
    \item \textbf{Understanding the Problem:} o1-preview demonstrated an improved ability to grasp the task requirements and constraints, particularly in well-defined, rule-based environments like \textit{Barman} and \textit{Tyreworld}. This was largely due to its self-evaluation mechanism, which allowed for more accurate state tracking and constraint adherence. However, more evidence is needed to establish whether these improvements translate to better reasoning capabilities in more abstract settings.
    
    \item \textbf{Following Constraints:} Across most tasks, o1-preview showed a superior capacity to follow task-specific constraints compared to GPT-4. However, this ability weakened as the complexity of spatial reasoning and state transitions increased, as seen in \textit{Termes}. This suggests that while constraint following is a relative strength of the o1 model, more work is needed to handle environments with higher-dimensional state spaces.
    
    \item \textbf{State and Memory Management:} One of o1-preview’s key advantages over previous models is its ability to remember and manage multiple states effectively within a plan, which contributed to its higher success rate in certain tasks. However, as problem complexity increased, the model’s state management became less reliable, particularly in tasks involving spatial reasoning across multiple dimensions. This implies a potential bottleneck in the model's memory and decision-making processes.
    
    \item \textbf{Reasoning and Generalization:} While o1-preview showed some promise in its generalization ability, particularly in structured environments like \textit{Grippers}, its performance in more abstract tasks like \textit{Termes} revealed substantial limitations. The model struggled with reasoning under conditions where actions and outcomes were less directly tied to the natural language representation of the task, highlighting an area for future improvements.
\end{enumerate}

\subsection{Opportunities for Improvement}
We posit several key areas where future iterations of LLM-based planners can be improved:

\begin{itemize}
    \item \textbf{Optimality and Resource Utilization:} Developing more sophisticated decision-making mechanisms that minimize redundant actions and optimize resource usage will be crucial for making o1 models more applicable to real-world planning tasks. This could involve incorporating cost-based reasoning or learning from expert demonstrations to achieve more optimal plans.
    
    Additionally, Retrieval-Augmented Generation (RAG) methods could offer a potential solution by providing real-time, low-cost external memory updates, especially when tasks rely on large knowledge bases encoded in natural language text. However, RAG’s effectiveness hinges on the accuracy and efficiency of its retrieval algorithms, which may introduce further challenges.

    \item \textbf{Generalization in Abstract Spaces:} While o1-preview shows promise in generalizing across structured environments, its performance in tasks with more abstract and complex rule sets remains suboptimal. Future work should focus on enhancing the model’s ability to generalize in high-dimensional and spatially dynamic environments, potentially through improved memory management \cite{yang2024text} and abstraction mechanisms \cite{zheng2024symbolic}.

    Enhancing the model’s decision-making and memory management capabilities, particularly for spatially complex tasks, will be essential for improving both optimality and generalizability in future iterations of LLM-based planning models.

    \item \textbf{Handling Dynamic and Unpredictable Environments:} Many real-world planning problems involve dynamic environments with unpredictable elements. Testing the o1 models in such settings would provide valuable insights into their robustness and adaptability, especially when rules or constraints change during execution.

    \item \textbf{Improving Constraint Adherence through Self-Evaluation:} One recurring issue across multiple domains is the models’ inability to follow task-specific constraints accurately. Introducing more robust self-evaluation mechanisms could help LLMs better verify their own outputs before finalizing decisions, potentially catching mistakes like rule violations. Techniques such as multi-stage validation or symbolic verification \cite{yang2024fine}, where models cross-check their proposed actions against the task constraints, could significantly reduce the incidence of constraint-related errors.

    \item \textbf{Leveraging Multimodal Inputs:} To enhance the model’s understanding of spatial and physical reasoning tasks, future LLM-based planners could benefit from integrating multimodal inputs such as visual data, 3D environments, or sensor information \cite{sun2024mm3dgs}. By incorporating non-textual data, planners would be better equipped to handle complex tasks, such as robotic manipulation or navigation, where purely text-based reasoning might miss critical spatial relationships or physical constraints.

    \item \textbf{Scalability to Complex Multi-Agent Planning:} Many planning tasks, particularly in robotics and logistics, require coordination between multiple agents \cite{wu2023autogen}. Extending LLM-based planners to effectively handle multi-agent systems would be an important step forward. This could involve developing strategies for decentralized planning, where each agent generates its own plan based on local knowledge, while still cooperating to achieve a shared goal.

    \item \textbf{Incorporating Human Feedback for Continuous Learning:} One way to improve both optimality and generalization is by incorporating continuous learning through human feedback. Interactive feedback loops, where human users provide corrective signals or suggestions during plan execution, could help models refine their decision-making and better adapt to new situations or tasks that deviate from their training data.

\end{itemize}

In conclusion, while o1-preview represents a notable advancement in LLM-based planning, significant challenges remain, particularly in terms of optimizing plans, generalizing to more abstract tasks, and managing state complexity. Future research should aim to build on these insights to create more robust, efficient, and adaptable planning agents capable of handling the diverse range of challenges presented by real-world planning problems.

\bibliographystyle{unsrtnat}
\bibliography{ref}

\end{document}